\documentclass[letterpaper, 10 pt, conference]{ieeeconf} 
\IEEEoverridecommandlockouts                             
\usepackage{algorithmic}
\usepackage[ruled,vlined,noend, noline]{algorithm2e}
\usepackage{amsfonts}
\usepackage{amsmath}
\usepackage{amssymb}
\usepackage{anyfontsize}
\usepackage[english]{babel}
\usepackage{booktabs}
\usepackage{caption}
\usepackage{comment}
\usepackage[keeplastbox]{flushend}
\usepackage[T1]{fontenc}
\usepackage[hidelinks]{hyperref}
\usepackage[utf8]{inputenc}
\usepackage{lipsum} 
\usepackage{mathtools}
\usepackage{microtype}
\usepackage{nicefrac}
\usepackage{setspace}
\usepackage[separate-uncertainty=true,detect-mode=true]{siunitx}
\usepackage{tabu}
\usepackage{url}

\newcommand{\smallsqueeze}{\vspace{-2.5mm}}

\usepackage[style=ieee,backend=bibtex,url=false,doi=false,natbib=true,mincitenames=1,maxcitenames=3,dashed=false,isbn=false]{biblatex}
\usepackage[capitalize]{cleveref}
\AtEveryBibitem{\clearfield{issn}}
\AtEveryCitekey{\clearfield{issn}}

\usepackage{svg}
\usepackage{color}
\usepackage{graphics}
\usepackage{pgfplots}
\usepackage{subcaption}
\pgfplotsset{compat=newest}
\usepackage{tikz}
\usetikzlibrary{shapes,arrows}
\definecolor{vandeusen}{RGB}{73,92,111}
\definecolor{cordovan}{RGB}{152,68,71}
\definecolor{middle_green}{RGB}{91,140,90}
\definecolor{honey}{RGB}{232, 174, 104}
\definecolor{vandeusen_light}{RGB}{128,141,154}

\definecolor{pastelPurple}{HTML}{8770FE} 
\definecolor{pastelBlue}{RGB}{0,114,178} 
\definecolor{pastelGreen}{RGB}{0,158,115} 
\definecolor{pastelRed}{HTML}{F5615C} 

\pgfplotsset{every axis legend/.append style={legend cell align=left}}
\usepgfplotslibrary{fillbetween}

\title{\LARGE \bf
Uncertainty-Aware Online Merge Planning with Learned Driver Behavior
}

\author{Liam A. Kruse, Esen Yel, Ransalu Senanayake, and Mykel J. Kochenderfer
\thanks{L. A. Kruse, E. Yel, R. Senanayake, and M. J. Kochenderfer are with the Stanford Intelligent Systems Laboratory in the Department of Aeronautics and Astronautics at Stanford University, Stanford, CA 94305, USA (email: \texttt{\{lkruse, esenyel, ransalu,  mykel\}@stanford.edu}).}
}

\begin{document}
\maketitle
\thispagestyle{empty}
\pagestyle{empty}

\begin{abstract}
Safe and reliable autonomy solutions are a critical component of next-generation intelligent transportation systems. Autonomous vehicles in such systems must reason about complex and dynamic driving scenes in real time and anticipate the behavior of nearby drivers. Human driving behavior is highly nuanced and specific to individual traffic participants. For example, drivers might display cooperative or non-cooperative behaviors in the presence of merging vehicles. These behaviors must be estimated and incorporated in the planning process for safe and efficient driving. In this work, we present a framework for estimating the cooperation level of drivers on a freeway and plan merging maneuvers with the drivers' latent behaviors explicitly modeled. The latent parameter estimation problem is solved using a particle filter to approximate the probability distribution over the cooperation level. A partially observable Markov decision process (POMDP) that includes the latent state estimate is solved online to extract a policy for a merging vehicle. We evaluate our method in a high-fidelity automotive simulator against methods that are agnostic to latent states or rely on \textit{a priori} assumptions about actor behavior. 

\end{abstract}

\section{Introduction}
Ensuring safe interactions between automated and human traffic participants is a crucial step to making autonomous driving in urban environments a reality. Autonomous vehicles (AVs) must reason about the evolution of highly stochastic driving scenarios to provide protection for human passengers. Predicting the behavior of other traffic participants is an essential step to modeling scene evolution. However, forecasting human behaviors is a challenging task, as myriad psychological and physiological factors determine a traffic participant's driving tendencies \cite{bhattacharyya2020online}. Furthermore, urban driving scenarios induce complex multi-agent interactions as traffic participants seek to complete individual objectives. For example, vehicles attempting to merge on a freeway must identify a gap in the traffic, while vehicles on the freeway may exhibit cooperative or non-cooperative behavior depending on their own latent characteristics. Consequently, reasoning about the intentions and future actions of agents is important for multi-agent driving interactions such as merging \cite{bouton2019cooperation}, intersection navigation \cite{bouton2017belief}, and freeway driving \cite{sunberg2017value}.

\begin{figure}[t]
\centering
\include{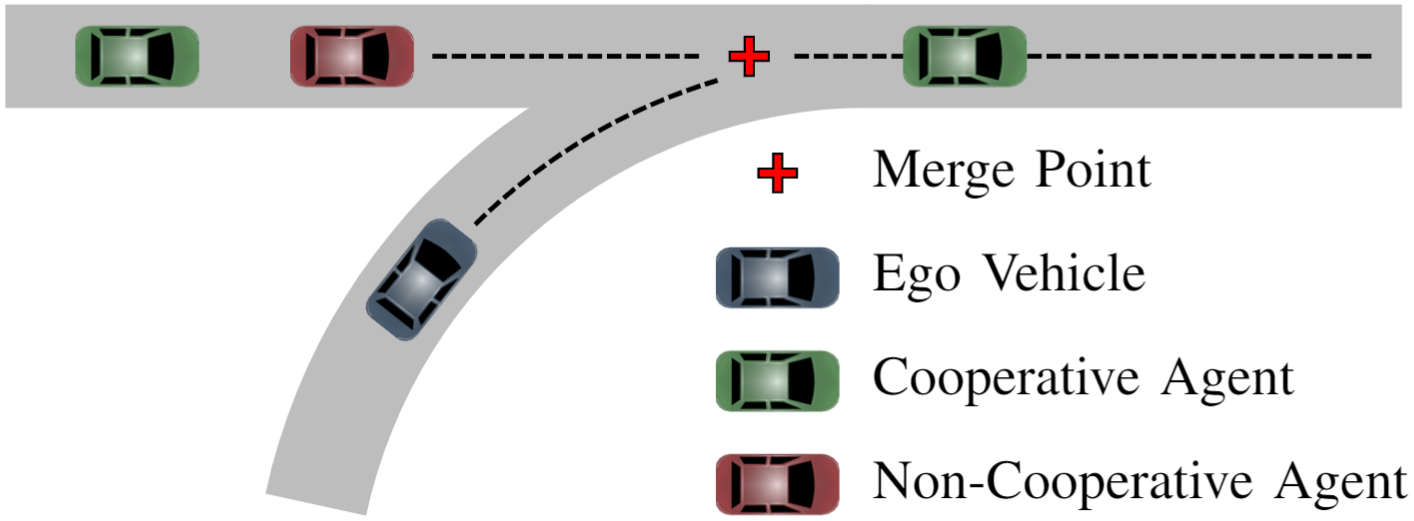}
\label{fig:merge_scene}
\end{figure}


Human driving behavior can be modeled using \textit{data-driven} or \textit{rule-based} methods. Data-driven models are trained on expansive driving datasets and are highly parameterized to capture nuanced driver characteristics. For example, generative adversarial networks and imitation learning techniques have been used to construct expressive models of human driving behavior \cite{bhattacharyya2020modeling}. However, data-driven approaches are \textit{black-box} in nature and difficult to interpret. Furthermore, data-driven models might induce dangerous behavior in driving scenes that do not appear in the training dataset \cite{bhattacharyya2020online}. An alternative to data-driven models is rule-based driver models such as the Intelligent Driver Model \cite{treiber2000congested}. Rather than extracting driving characteristics from data, rule-based models specify a mathematical model using a set of rules derived from expert knowledge. Although such models rely on assumptions of human driving characteristics and can fail to generalize across a range of scenarios, they are highly interpretable and provide useful insights into microscopic human driving behavior. \citet{brown2020taxonomy} present an extensive survey of human driver models.

Once the parameters of a driver model have been specified, the behavior of traffic participants can be forecasted and input to a decision-making system such as open-loop planing, or \textit{model predictive control} \cite{brown2017safe, leung2018infusing}. Game-theoretic frameworks \cite{wang2021game} and reinforcement learning \cite{ma2021reinforcement, bouton2019cooperation} have also been used for automotive decision making, but these strategies are computationally expensive. Partially observable Markov decision processes (POMDPs) are well-suited for AV decision making because they account for uncertainty in the environment and latent behaviors of traffic participants, and have been used with great success in the freeway driving context \cite{sunberg2017value, sunberg2020improving}. In this work, we model the merge scenario shown in \cref{fig:merge_scene} as a POMDP and solve the POMDP online for a driving policy. \textit{Online} driving strategies choose the next action using real-time sensor information, while \textit{offline} driving strategies plan all actions in advance. 

Several papers have investigated interaction-aware planning for merge scenarios. \citet{ward2017probabilistic} considered merging on freeways and T-junctions; they fit parameters to an extension of the IDM using a nonlinear least-squares procedure. \citet{bouton2019cooperation} and \citet{ma2021reinforcement} conducted reinforcement learning-based techniques to learn merging policies. These works represent offline strategies for parameter estimation; we instead pursue an online strategy to capture the idiosyncrasies of individual drivers in real-time. The strategy presented by \citet{sunberg2020improving} is most similar to our approach; however, they only consider lane change merging. We investigate freeway merging, which has been qualitatively described as one of the riskiest and complex driving maneuvers frequently encountered in day-to-day driving \cite{ward2017probabilistic}.

In this paper, we perform online parameter estimation for a rule-based driver model---the Cooperative Intelligent Driver Model (C-IDM) \cite{bouton2019cooperation}---that explicitly models the cooperation level of vehicles on a main freeway lane encountering a merging vehicle. The estimated cooperation level informs an online POMDP-based planner that executes a merge maneuver with awareness of latent driver behaviors. Our specific contributions are:
\begin{itemize}
\item We formulate the latent behavior prediction task as a recursive Bayesian estimation problem and estimate the cooperation level of vehicles using a particle filter.
\item We use the estimates of latent driver behaviors to inform an online POMDP-based planner which controls the merging vehicle. We thus account for two levels of uncertainty, as we propagate uncertainty over latent behavior through the planning framework.
\item We evaluate the safety and efficiency of our driving strategy in a high-fidelity automotive simulator and benchmark our planning strategy against methods that do not model latent behaviors or rely on \textit{a priori} assumptions about agent behavior. 
\end{itemize}

\section{Driver Models}

Rule-based driver models can provide reasonable approximations of microscopic driver behavior with a relatively small number of parameters, making them ideally suited to the online parameter estimation framework. In this section we discuss the ubiquitous IDM \cite{treiber2000congested} and an IDM extension for merge scenarios, the Cooperative IDM \cite{bouton2019cooperation}. We also present approaches for learning driver model parameters.

\subsection{Intelligent Driver Model}
The Intelligent Driver Model is an adaptive cruise control model that governs longitudinal vehicle motion. The IDM specifies a mathematical model that balances an agent's desire to travel at a particular speed with the need to maintain adequate separation distance between a leader vehicle. 
Given a current speed $v(t)$, a distance headway $d(t)$, and a desired speed $v_{\mathrm{des}}$, the IDM computes a desired acceleration according to

\begin{equation}
    {a_{{\mathrm{IDM}}}} = {a_{\max }}\left( {1 - {{\left( {\frac{{v(t)}}{{{v_{{\mathrm{des}}}}}}} \right)}^4} - {{\left( {\frac{{{d_{{\mathrm{des}}}}}}{{d(t)}}} \right)}^2}} \right).
\end{equation}
The desired separation distance $d_{\mathrm{des}}$ is defined as
\begin{equation}
    {d_{{\mathrm{des}}}} = {d_{{\mathrm{min}}}} + \tau \cdot v(t) - \frac{{v(t) \cdot r(t)}}{{2\sqrt {{a_{{\mathrm{max}}}} \cdot {b_{{\mathrm{pref}}}}} }}.
\end{equation}

Model inputs include the minimum allowable separation distance $d_{\mathrm{min}}$; the minimum allowable separation time $\tau$; the acceleration limit $a_{\mathrm{max}}$; and the deceleration limit $b_{\mathrm{pref}}$. 

\subsection{Cooperative Intelligent Driver Model}

The Cooperative Intelligent Driver Model is an extension of the IDM developed by \citet{bouton2019cooperation} for merge scenarios. The model governs the longitudinal acceleration of a vehicle on the main lane while considering the time to reach the merge point ($TTM$) for a merging agent. The C-IDM is parameterized by the IDM parameters as well as the \textit{cooperation level} $c \in [0,1]$ that determines the extent to which the driver cooperates with the merging vehicle. A cooperation level of $c = 1$ corresponds to a \textit{cooperative} agent that yields to the merging vehicle, while a cooperation level of $c = 0$ corresponds to a \textit{non-cooperative} agent that ignores the merging vehicle. 

Consider the merge scenario shown in \cref{fig:cidm_setup} with an ego vehicle $V_{\mathrm{E}}$, a leading agent on the main lane $V_{\mathrm{L}}$, and a trailing agent on the main lane $V_{\mathrm{T}}$. Let $TTM(V_{i})$ represent the time required for the $i$th actor to reach the specified merge point. C-IDM outputs the longitudinal acceleration for the trailing agent according to the following rule:

\begin{itemize}
    \item If $TTM(V_{\mathrm{E}}) < c \times TTM(V_{\mathrm{T}})$, then $V_{\mathrm{T}}$ follows IDM while considering the projection of the ego on the main lane as its leading vehicle.
    \item If $TTM(V_{\mathrm{E}}) \geq c \times TTM(V_{\mathrm{T}})$ or if no merging vehicle exists, then $V_{\mathrm{T}}$ follows IDM while considering the next actor on the main lane, $V_{\mathrm{L}}$, as its leading vehicle.
\end{itemize}
\begin{figure}[h!]
\centering
\include{figs/cidm_setup}
\label{fig:cidm_setup}
\end{figure}

\subsection{Rule-Based Driver Model Parameter Estimation}

Rule-based driver models are governed by a relatively small number of parameters, which can be estimated in either an \textit{offline} or an \textit{online} fashion. Offline methods typically aggregate driving data in the training dataset and thus lose information on individual traffic participants. However, such methods can use arbitrarily large computation times and dataset sizes to maximize accuracy. Conversely, online methods must operate with constraints on computation time and data availability but capture the behavior of individual drivers. Offline strategies for rule-based driver model parameter estimation include constrained nonlinear optimization \cite{lefevre2014comparison}, the Levenberg-Marquardt algorithm \cite{morton2016analysis}, and genetic algorithms \cite{kesting2008calibrating}. Online strategies include recursive Bayesian filters such as the Extended Kalman filter \cite{monteil2015real} and the particle filter \cite{bhattacharyya2021hybrid, buyer2019interaction, schulz2018interaction}. We employ a particle filter to conduct latent parameter estimation due to its ability to represent complex belief states.
\section{Methodology}

We model the merge scenario as a partially observable Markov decision process, which can be defined by the tuple $(\mathcal{S},\mathcal{A},T,R,\mathcal{O},Z,\gamma)$, with $\mathcal{S}$ the state space, $\mathcal{A}$ the action space, $T$ the transition model, $R$ the reward function, $\mathcal{O}$ the observation space, $Z$ the observation likelihood function, and $\gamma$ the discount factor. An agent in state $s \in \mathcal{S}$ takes action $a \in \mathcal{A}$ and transitions to a successor state $s^{\prime}$ according to the transition model $T(s^{\prime},s,a) = \mathrm{Pr}(s^{\prime} \mid s, a)$. The agent receives a reward according to the reward function $R(s,a,s^{\prime})$. Rather than observing the true state, the agent receives a potentially imperfect observation $o \in \mathcal{O}$ at each time step. The likelihood of receiving observation $o$ is governed by $Z(o,s,a) = \mathrm{Pr}(o \mid s, a)$. Finally, the discount factor $\gamma$ can be adjusted to make the agent more or less myopic. We model the merge scenario with the following definitions:

\subsection{The State Space}

The state of a given driving scene,
\begin{equation*} s=(V_{E},\ V_{T},\ c_{T},\ V_{L})\in \mathcal{S}, \end{equation*}
consists of the observable states $V_{E}$, $V_{T}$, $V_{L}$ of the ego, trailing agent, and lead agent, respectively, and the latent cooperation level $c_{T}$ which governs the C-IDM for the trailing agent. The observable vehicle state
\begin{equation*} V_{i}=(x_{i},\ y_{i},\ v_{i},\ \dot{v}_{i},\ \theta_{i}) \end{equation*}
consists of the $i$th vehicle's position $(x_{i},\ y_{i})$ on the simulator map, longitudinal velocity $v_{i}$, longitudinal acceleration $\dot{v}_{i}$, and heading angle $\theta_{i}$.

\subsection{The Action Space}

At each time step, the ego takes an action $a \in \mathcal{A}$ which consists of a longitudinal jerk value. Planning occurs in jerk space to produce smooth velocity profiles. Jerk values are restricted to the range $\SI{-0.6}{\meter/\second^3} \leq a \leq \SI{0.6}{\meter/\second^3}$ to prevent unrealistic or unsafe motion.

\subsection{The Transition Model}

The transition model is implicitly defined by a generative model, $s^{\prime} \sim T(s,a)$, which generates successor states from the distribution  $\mathrm{Pr}(s^{\prime} \mid s, a)$. Vehicles on the main lane propagate according to one-dimensional point mass dynamics:
\begin{align*} & {v^{t + 1}} = {v^t} + {\dot{v}^t}\Delta t \\ & {x^{t + 1}} = {x^t} + {v^t}\Delta t + \frac{1}{2}{\dot{v}^t}\Delta {t^2} \end{align*}

Lateral speed is assumed to be zero. The trailing agent's acceleration is governed by the C-IDM, while the lead agent moves with constant velocity. The ego vehicle propagates according to a state transition function 
\begin{align*}  & {\dot{v}^{t + 1}} = {\dot{v}^t} + \ddot{v}^t\Delta t \\ & {v^{t + 1}} = {v^t} + {\dot{v}^t}\Delta t 
\end{align*}
The applied jerk $\ddot{v}^t$ is the action $a$ extracted at the current time step. We assume that the ego travels a distance ${v^t}\Delta t$ along the merge lane curve during each discrete time step.

\subsection{The Reward Function}

Autonomous vehicles must balance multiple and occasionally competing objectives such as safety, control effort, and rider comfort. We construct the following reward function to promote safe and efficient merge maneuvers:
\begin{equation} R(s)= -\lambda_1 \Vert v_{E} - v_{\text{ref}}\Vert - \lambda_2 \Vert \dot{v}_{E} \Vert  - \lambda_3 \mathbf{1}_{b_{\text{hard}}} \end{equation}
where $v_{\text{ref}}$ is the ego vehicle's desired reference velocity and $\mathbf{1}_{b_{\text{hard}}}$ is an indicator function of hard braking:
\begin{equation*}
\mathbf{1}_{b_{\text{hard}}} = \begin{cases}
{1}&{\text{if}\;\text{$\Vert(x_E,\ y_E) - (x_T,\ y_T) \Vert < d_{\mathrm{safety}} $}}\\
{0}&{\text{otherwise}}\\
\end{cases}
\end{equation*}
where $d_{\mathrm{safety}}$ is a user-defined safety threshold. We define \textit{hard braking} as an uncomfortably abrupt deceleration that can be applied to prevent a collision. The hard brake rate is thus a proxy metric for AV safety and rider comfort across simulations. The weights $\lambda_i$ can be adjusted to modify the ego's preferences for safety, efficiency, and control effort. 

\subsection{Online Parameter Estimation}


Given a sequence of observations of the trailing agent's position, we wish to obtain a distribution over the latent cooperation level $c_T$. We assume that the cooperation level is time-invariant during the brief interaction between the trailing agent and the merging ego. Our proposed approach frames the task of estimating latent driver behavior as a state estimation problem and solves the inference problem using recursive Bayesian estimation. The general form of the recursive Bayesian update equation is 
\begin{equation}
    p(\theta_T \mid \mathbf{y}_{1:t}) = \frac{p(\mathbf{y}_t \mid \theta_T)p(\theta_T \mid \mathbf{y}_{1:t-1})}
    {\int_{\theta_T} p(\mathbf{y}_t \mid \theta_T)p(\theta_T \mid \mathbf{y}_{1:t-1}) \mathrm{d}\theta_T} \text{.}
    \label{eqn:recursive_state_estimation}
\end{equation}

Because \cref{eqn:recursive_state_estimation} cannot be solved analytically \cite{bhattacharyya2020online}, we use particle filtering to provide an approximate solution to the inference problem. \Cref{alg:particle_filtering} presents our strategy for estimating the cooperation level of the trailing agent. The particles are initially sampled from a uniform distribution and are then propagated according to the POMDP transition model. To prevent particle deprivation, we add noise $d_{c_T}$ sampled from a discrete uniform distribution $D = \mathcal{U}(\{-0.05,\ 0,\ 0.05 \})$.

\begin{algorithm}[htb]

\setstretch{1.2}

\textbf{Input:} Initial particle set $\Theta$, ego $V_E$, trailing agent $V_T$, leading agent $V_L$, POMDP model $\mathcal{P}$ \\
$c_{T}^{0} \leftarrow 0.5$ (initialize cooperation parameter)\\
\For{$t \gets 0, 1, \ldots, N\ \mathrm{(time \ steps)}$}{
    $x_T^{t+1} \leftarrow \mathrm{true \ position \ of \ trailing \ agent}$\\
    \For{$i \leftarrow 1, 2, \ldots, K\ \mathrm{(particles)}$}{
        \uIf{$TTM(V_{\mathrm{E}}) < c \times TTM(V_{\mathrm{T}})$}{
            $\dot{v}_{T}^{t+1} \leftarrow$ IDM with ego projection as leader\
        }
        \Else{
            $\dot{v}_{T}^{t+1} \leftarrow$ IDM with lead agent as leader\
        }
        Propagate vehicles according to $\mathcal{P}.T(s^{\prime}, a, s)$ \\
        $w_{i} \gets Z\left(x_T^{t+1} \ | \ x_{T,i}^{t+1}\right)$\\
        $d_{c_T} \sim D$ (dithering)\\
        $c_{T}^{t+1} \leftarrow c_{T}^{t} + d_{c_{T}}$, clamped between $0$ and $1$
    }
    $\Theta \leftarrow \mathrm{Resample \ from \ } \Theta \propto \left[ w_{1}, \ldots, w_{K}\right]$
}\caption{Online Parameter Estimation}
\label{alg:particle_filtering}
\end{algorithm}
\section{Experiments}

In this work, we consider a freeway merge scenario as visualized in \cref{fig:cidm_setup}. An ego vehicle---representing either an autonomous agent or a human-driven vehicle with a guardian software system---seeks a gap on the main freeway lane. The vehicles are positioned such that the projection of the ego vehicle is in front of the trailing agent. Furthermore, we induce a conflict between the ego and the trailing agent by selecting initial velocities that place them on an intercept trajectory at the merge point. \Cref{tab:experimental_setup} shows key values for the POMDP reward functions. The POMDP was solved online using partially observable Monte Carlo planning with observation widening (POMCPOW) \cite{sunberg2018online}.

\begin{table}\caption{\label{tab:experimental_setup} POMDP Reward Function Parameters}
\centering
\begin{tabular}{@{}lr@{}} \toprule
    \textbf{Parameter} & \textbf{Value}  \\
    \midrule
    $\lambda_1$ (speed deviation) & $1$ \\
    $\lambda_2$ (control effort)& $1$ \\
    $\lambda_3$ (collision) & $100$ \\
    $d_{\mathrm{safety}}$ & $\SI{15.0}{\meter}$ \\\bottomrule
\end{tabular}
\smallsqueeze
\smallsqueeze
\end{table}

\subsection{Benchmark Methods}
We tested our approach for learning the cooperation parameter against two other POMDP-based planning approaches that make \textit{a priori} assumptions about the trailing agent's cooperation level: first, a cautious approach where the trailing agent is always assumed to be non-cooperative (e.g., the ego always assumes $c = 0.0$), and second, a risky approach where the trailing agent is always assumed to be cooperative (e.g., the ego always assumes $c = 1.0$). We also benchmarked against a scenario wherein the ego acceleration was governed by the stochastic IDM (SIDM)---an extension of the IDM that outputs noisy acceleration values \cite{treiber2017intelligent}. We reproduced the merge scene shown in \cref{fig:merge_scene} using the Applied Intuition simulation engine and ran 500 trials for each driving strategy; the trailing agent was assigned a ground truth cooperation level of $c = 0.0$ in 250 trials and a ground truth cooperation level of $c = 1.0$ in the other 250 trials.
\subsection{Filtering}
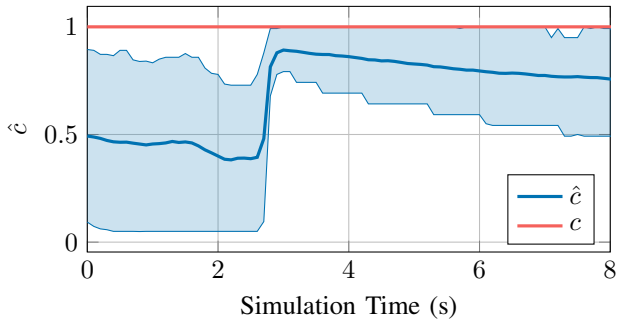
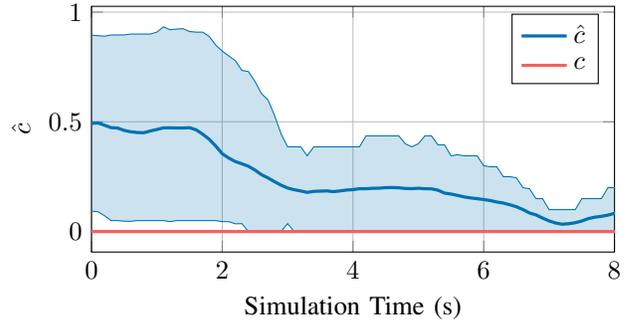
\begin{figure*}[h]
    \begin{subfigure}[b]{0.48\textwidth}
       \centering
       \begin{tikzpicture}[scale = 1.0]
\begin{axis}[
  ylabel = {$\hat{c}$},
  legend pos = {south east},
  xmin = {0.0},
  xmax = {8.0},
  xlabel = {Simulation Time (s)},
  enlarge x limits=false,grid=both,
  width=\textwidth,
  height=4.85cm
]

\addplot [name path = A, pastelBlue, forget plot] table [x = {time}, y = {minimum}, col sep=comma] {figs/coop_new.csv};

\addplot [name path = B, pastelBlue, forget plot] table [x = {time}, y = {maximum}, col sep=comma] {figs/coop_new.csv};

\addplot [pastelBlue, very thick] table [x = {time}, y = {mu}, col sep=comma] {figs/coop_new.csv}; \addlegendentry{$\hat{c}$};

\addplot [pastelBlue,opacity = 0.2, forget plot] fill between [of = A and B];

\addplot [mark=none, pastelRed, very thick, domain=0:8.2] {1.0}; \addlegendentry{$c$};

\end{axis}
\end{tikzpicture}
        \caption{Sample cooperation level estimation when the trailing agent is fully cooperative, e.g., $c = 1.0$}
        \label{fig:est_coop}
    \end{subfigure}
    \hfill
    \begin{subfigure}[b]{0.48\textwidth}
        \centering
        \begin{tikzpicture}[scale = 1.0]
\begin{axis}[
  ylabel = {$\hat{c}$},
  legend pos = {north east},
  xmin = {0.0},
  xmax = {8.0},
  xlabel = {Simulation Time (s)},
  enlarge x limits=false,grid=both,
  width=\textwidth,
  height=4.85cm
]

\addplot [name path = A, pastelBlue, forget plot] table [x = {time}, y = {minimum}, col sep=comma] {figs/noncoop_new.csv};

\addplot [name path = B, pastelBlue, forget plot] table [x = {time}, y = {maximum}, col sep=comma] {figs/noncoop_new.csv};

\addplot [pastelBlue, very thick] table [x = {time}, y = {mu}, col sep=comma] {figs/noncoop_new.csv}; \addlegendentry{$\hat{c}$};

\addplot [pastelBlue,opacity = 0.2, forget plot] fill between [of = A and B];

\addplot [mark=none, pastelRed, very thick, domain=0:8.2] {0.0}; \addlegendentry{$c$};

\end{axis}
\end{tikzpicture}
        \caption{Sample cooperation level estimation when the trailing agent is fully non-cooperative, e.g., $c = 0.0$}
        \label{fig:est_noncoop}
    \end{subfigure}
    \hfill
\caption{Examples of online latent parameter estimation for the trailing agent's cooperation level. The estimate of $c$ quickly approaches the true value, indicating that the ego is aware of the trailing agent's cooperation level as it nears the merge point.}
\label{fig:parameter_estimation}
\smallsqueeze
\end{figure*}

\Cref{fig:parameter_estimation} shows two examples of cooperation level parameter estimation drawn from the batch of experiments. The red line indicates the true cooperation level of the trailing agent while the blue line represents the ego's estimate of the cooperation level. The light blue regions show the 90\% empirical error bounds. The parameter estimate is initialized to $c=0.5$ with diffuse error bounds. The ego vehicle observes the trailing agent either slow down to facilitate a merge (\cref{fig:est_coop}) or maintain its current speed while ignoring the ego (\cref{fig:est_noncoop}) and updates its estimate of the cooperation level accordingly. \Cref{fig:parameter_estimation} indicates that the ego is able to quickly identify whether a trailing agent is behaving in a generally cooperative or uncooperative manner. The fact that the parameter estimates do not converge exactly to the true value is not unexpected; since the C-IDM relates the time to merge of the ego and the trailing agent with an inequality, there are in general a range of cooperation levels that will yield the same behavior from the trailing agent. A visualization of parameter estimation during merge scenarios was created using Applied Intuition simulation tools and can be found on the lab video channel at \url{https://youtu.be/KnM3azGH_Sg}.

\subsection{Performance Results}
We count the occurrences of hard braking and divide by the total number of trials to obtain the hard brake rate across all experimental configurations. We also measure \textit{time-to-collision} (TTC), which is a safety metric, and \textit{time-to-merge}, which indicates how efficient a merge maneuver is. 
Higher TTC values are associated with safer driving scenes \cite{minderhoud2001extended}, while excessively high time-to-merge values could be indicative of the \textit{freezing robot problem} wherein the robot freezes in place to avoid a collision \cite{trautman2010unfreezing}.

\Cref{tab:hard-brake-rate} displays the rate of hard brake incidents across all driving methods. Hard brake rates remain predictably low across experiments with a cooperative trailing agent since the trailing agent avoids conflict with the merging ego by considering its projection on the main lane. When the trailing agent is non-cooperative, an upper bound on performance is given by the POMDP-based planner that assumes full knowledge of the true cooperation level of $c = 0.0$. Our proposed method for learning $c$ and the strategy of fixing $c = 0.0$ complete all trials without a hard brake incident. Planning with a fixed $c = 1.0$ expresses overconfidence in the desire of the vehicle on the main lane to cooperate and results in a hard brake rate of over $65\%$. The SIDM driving strategy, which does not consider the cooperation level of the vehicle on the main lane, must conduct hard braking on every trial to avoid a collision.

\begin{figure*}[h]
    \begin{subfigure}[t]{0.48\textwidth}
    \centering
       \input{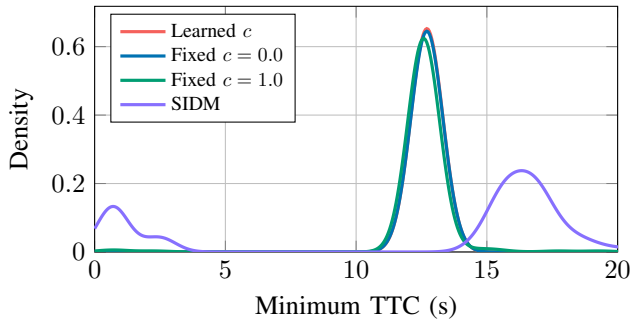}\caption{Distribution over minimum time-to-collision values achieved in trials with a cooperative agent.}
        \label{fig:min_ttc_a1}
    \end{subfigure}
    \hfill
    \begin{subfigure}[t]{0.48\textwidth}
    \centering
        \input{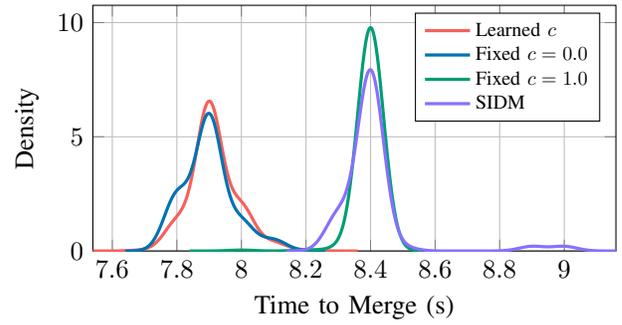}\caption{Distribution over time required to merge in trials with a cooperative agent.}
        \label{fig:ttm_a1}
    \end{subfigure}
    \hfill
\caption{Experimental results given a cooperative agent on the main lane. Our strategy for learning $c$ achieves comparable results to the nominal planner that assumes full knowledge of the trailing agent's true cooperation level of $c = 1.0$.}
\label{fig:metrics_a1}
\end{figure*}

\begin{figure*}[h]
    \begin{subfigure}[t]{0.48\textwidth}
    \centering
       \input{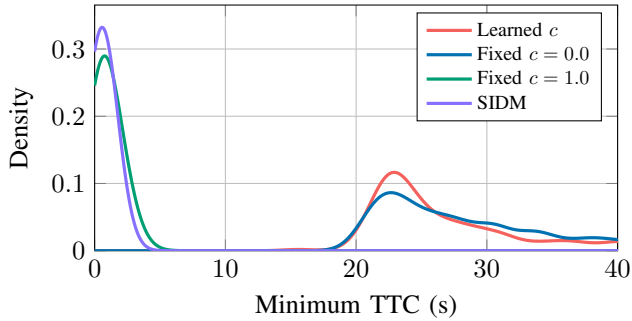}\caption{Distribution over minimum time-to-collision values achieved in trials with a non-cooperative agent.}
        \label{fig:min_ttc_a0}
    \end{subfigure}
    \hfill
    \begin{subfigure}[t]{0.48\textwidth}
    \centering
        \input{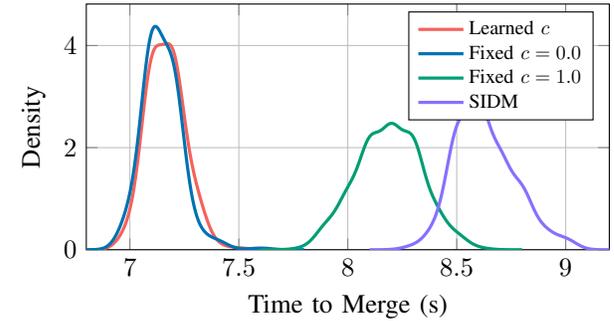}\caption{Distribution over time required to merge in trials with a non-cooperative agent.}
        \label{fig:ttm_a0}
    \end{subfigure}
    \hfill
\caption{Experimental results given a non-cooperative agent on the main lane. Our strategy for learning $c$ achieves comparable results to the nominal planner that assumes full knowledge of the trailing agent's true cooperation level of $c = 0.0$.}
\label{fig:metrics_a0}
\smallsqueeze
\end{figure*}

\begin{table}\caption{\label{tab:hard-brake-rate}Hard Brake Rate}
\centering
\begin{tabular}{@{}lcr@{}} \toprule
    \textbf{Driving Strategy} & \multicolumn{2}{c}{\textbf{Hard Brake Rate (\%)}}  \\
    \midrule
    & $c_T = 1.0$ & $c_T = 0.0$ \\
    \midrule
    Learned $c$ & $0.0$ & $0.0$\\ 
    Fixed $c = 0.0$ & $0.0$ & $0.0$ \\
    Fixed $c = 1.0$ & $0.4$ & $65.2$\\
    SIDM & $4.0$ & $100.0$\\\bottomrule
\end{tabular}
\smallsqueeze
\smallsqueeze
\end{table}
\Cref{fig:metrics_a1} presents the results for the trials with a cooperative trailing agent. The POMDP-based driving strategies achieve average TTC values greater than $10$ seconds. Many SIDM trials also achieve high TTC values; however, the SIDM trials that result in hard brake incidents are associated with small TTC values. Our strategy for learning $c$ achieves comparable results to the planner that assumes full knowledge of the trailing agent's true cooperation level of $c = 1.0$. Note that assuming non-cooperative behavior does not underperform in these trials since the scenario is designed to induce conflict. In fact, assuming a non-cooperative agent leads to smaller time-to-merge values as the ego accelerates to reach the gap ahead of the trailing agent, as shown in \cref{fig:ttm_a1}.
\Cref{fig:metrics_a0} displays the results for the trials with a non-cooperative trailing agent. Both the SIDM and fixed $c = 1.0$ methods suffered from elevated hard brake rates and correspondingly low average TTC values. Our proposed method and the nominal benchmark obtained by fixing $c = 0.0$ achieved higher TTC values as shown in \cref{fig:min_ttc_a0}. An explanation for the difference in TTC values is clear from \cref{fig:ttm_a0}. Our proposed approach and the nominal baseline experience decreased time-to-merge values as they accelerate into the gap in front of the trailing agent. The POMDP-based driving strategy that assumes cooperative behavior takes more time to merge and thus encounters a conflict with the trailing agent at the merge point. The SIDM approach does not consider the behavior of the trailing agent at all and thus takes the longest time to merge on average.

\section{Conclusion and Future Work}
This paper presented a method for conducting online merge planning with awareness of the latent behaviors of vehicles on the main lane. We modeled the behavior of agents on a freeway using the Cooperative Intelligent Driver Model and demonstrated accurate estimation of the latent cooperation level using a particle filter. Our estimate of the cooperation level informed a behavior-aware POMDP model that was solved online for a merging policy. The performance of the proposed strategy closely matched the results achieved by baseline strategies that assumed full knowledge of other agents' latent characteristics. Furthermore, our driving strategy greatly outperformed a rule-based driving strategy that did not consider the cooperation level of other agents. Although the application described in this paper is specific to freeway merge scenarios, the parameter estimation and online planning framework is quite general and can easily be extended to other driving scenarios or robotic applications.

This work assumed that the latent behavior of vehicles was time-invariant; future work will relax this assumption to consider agents that change their behavior during a driving scene. 
Furthermore, the safety and efficiency of a merging ego vehicle will be improved by quantifying the position and intention uncertainty of agents on a freeway. Uncertainty-aware models of freeway agents will inform the construction of probabilistically safe ego trajectory envelopes.

\section*{Acknowledgments}
Toyota Research Institute (TRI) provided funds to assist the authors with their research, but this article solely reflects the opinions and conclusions of its authors and not TRI or any other Toyota entity. This research was supported in part by the National Science Foundation Graduate Research Fellowship Program under Grant No. DGE-1656518. We thank Applied Intuition for support with the simulation platform.

\renewcommand*{\bibfont}{\footnotesize}
\printbibliography

\end{document}